\documentclass[conference]{template/IEEEtran}
\IEEEoverridecommandlockouts

\usepackage{times}
\usepackage{soul}
\usepackage{url}
\usepackage[hidelinks]{hyperref}
\usepackage[utf8]{inputenc}
\usepackage[small,compatibility=false]{caption}
\usepackage{graphicx}
\usepackage{amsmath}
\usepackage{amsthm}
\usepackage{booktabs}
\usepackage{algorithm}
\usepackage{algorithmic}
\usepackage[switch]{lineno}
\usepackage{array}
\usepackage{xcolor}
\usepackage{colortbl}
\usepackage{arydshln}
\usepackage{stfloats}
\usepackage{multirow}

\usepackage{latexsym}

\usepackage{cite}
\usepackage{amsmath,amssymb,amsfonts}
\usepackage{algorithmic}
\usepackage{graphicx}
\usepackage{textcomp}
\usepackage{xcolor}
\def\BibTeX{{\rm B\kern-.05em{\sc i\kern-.025em b}\kern-.08em
    T\kern-.1667em\lower.7ex\hbox{E}\kern-.125emX}}
\begin{document}

\title{GroundCount: Grounding Vision-Language Models with Object Detection for Mitigating Counting Hallucinations}

\author{
    Boyuan Chen\textsuperscript{1,2}, 
    Minghao Shao\textsuperscript{1,2}, 
    Siddharth Garg\textsuperscript{1}, 
    Ramesh Karri\textsuperscript{1},
    Muhammad Shafique\textsuperscript{2}, 
    \\
    \textsuperscript{1}Tandon School of Engineering, New York University, NY, USA \\
    \textsuperscript{2}eBRAIN Lab, Division of Engineering, New York University Abu Dhabi, UAE \\
    \texttt{\{boyuan.chen, minghao.shao, sg175, rkarri, muhammad.shafique\}@nyu.edu}
}

\maketitle

\begin{abstract}
Vision Language Models (VLMs) exhibit persistent hallucinations in counting tasks,
with accuracy substantially lower than other visual reasoning tasks (excluding sentiment). This phenomenon
persists even in state-of-the-art reasoning-capable VLMs. Conversely, CNN-based
object detection models (ODMs) such as YOLO excel at spatial localization and
instance counting with minimal computational overhead. We propose
\textit{GroundCount}, a framework that augments VLMs with explicit spatial grounding
from ODMs to mitigate counting hallucinations. In the best case, our prompt-based augmentation strategy achieves 81.3\% counting accuracy on the best-performing model (Ovis2.5-2B)  - a 6.6pp improvement - while reducing inference time by 22\% through elimination of hallucination-driven reasoning loops for stronger models. We conduct
comprehensive ablation studies demonstrating that positional encoding is a critical
component, being beneficial for stronger models but detrimental for weaker ones.
Confidence scores, by contrast, introduce noise for most architectures and their
removal improves performance in four of five evaluated models. We further evaluate
feature-level fusion architectures, finding that explicit symbolic grounding via
structured prompts outperforms implicit feature fusion despite sophisticated
cross-attention mechanisms. Our approach yields consistent improvements across
four of five evaluated VLM architectures (6.2--7.5pp), with one architecture
exhibiting degraded performance due to incompatibility between its iterative
reflection mechanisms and structured prompts. These results suggest that counting
failures stem from fundamental spatial-semantic integration limitations rather
than architecture-specific deficiencies, while highlighting the importance of
architectural compatibility in augmentation strategies.
\end{abstract}

\section{Introduction}
\label{sec:intro}
Vision-Language Models (VLMs) have demonstrated remarkable progress across a wide range of multimodal tasks \cite{shao2024survey}, benefiting from large-scale training and fine-tuning on massive image-text corpora. This joint learning paradigm significantly enhances both contextual reasoning and instruction-following abilities \cite{liu2024llavanext}. Despite such advances, VLMs exhibit \textit{hallucinations} by generating object attributes, entities, or spatial relations inconsistent with the visual input \cite{zhao2025mitigating, zou2025memvr}. 

A growing body of work attributes this issue to an imbalanced cross-modal attention mechanism: the language decoder often over-attends to textual priors while under-utilizing visual tokens \cite{Leng_2024_CVPR, wang-etal-2024-mitigating, chuang2024dola}. Consequently, several methods have been proposed to mitigate hallucination, either through decoding-level adjustments \cite{xu-etal-2025-mitigating, yin2025clearsight, jiang2025devils} or training-level regularization \cite{wu2025antidote, yang2025mitigating}. Motivated by layer-level analysis showing that hallucination primarily occurs when later layers deviate from correct logits, another direction proposes steering strategies \cite{li2025the, wang2025damo, wang2025mllm} that preserve the correct momentum from earlier layers. These approaches have yielded measurable improvements on popular VLMs, including LLaVA-v1.5 \cite{liu2023llava}, LLaVA-v1.6 \cite{liu2024llavanext}, InstructBLIP \cite{dai2023instructblipgeneralpurposevisionlanguagemodels}, MiniGPT-2 \cite{chen2023minigptv2largelanguagemodel}, and mPLUG-Owl2 \cite{Ye2024mPLUG-Owl2}. 

Recent work jointly trains vision encoders and language decoders on diverse multimodal datasets \cite{zhu2025internvl3}, and incorporates reinforcement learning to enhance reasoning capabilities \cite{yang2025r4b, wang2025internvl35, lu2025ovis25}. These efforts have led to significantly stronger multimodal representations and improved factual alignment. However, even state-of-the-art (sota) VLMs continue to exhibit systematic hallucination in \emph{counting} tasks. As we demonstrate in Section~\ref{sec:analysis} and Table~\ref{tab:results_baseline}, counting consistently remains the lowest-accuracy task across all evaluated VLMs (excluding sentiment, which is relatively subjective), with accuracies ranging from 64.0\% to 74.7\%. This is substantially lower than other visual reasoning tasks such as object recognition (70.3\% to 89.6\%) and attribute identification (83.2\% to 86.5\%). Recent research also supports this finding \cite{vo2025vision, guo2025visionlanguagemodelcantcount}.

The persistence of counting hallucinations in reasoning-capable VLMs presents a unique challenge. Unlike earlier VLMs where attention steering and decoding adjustments proved effective, modern reasoning VLMs already incorporate reflection mechanisms \cite{lu2025ovis25, zhu2025internvl3, wang2025internvl35} that direct attention to visual modalities. Furthermore, the iterative reasoning process makes it difficult to apply layer-level vector steering, as there is no single "correct" token to steer toward during multi-step reasoning. This suggests that existing hallucination mitigation strategies, while effective for simpler VLMs, are insufficient for addressing systematic failures in compositional tasks like counting.

On the other hand, counting the number of occurrences of an object is nearly trivial for object detection models (ODMs), such as YOLO \cite{lei2025yolov13} and DETR \cite{carion2020detr}. These models achieve very high accuracy in object detection while requiring minimal inference time compared to auto-regressive VLMs. Moreover, ODMs provide structured outputs including bounding boxes, confidence scores, and class labels. This information explicitly addresses the spatial and compositional reasoning needed for accurate counting. This observation suggests a complementary approach: rather than attempting to fix VLM attention mechanisms or reasoning processes, we can augment VLMs with explicit grounding information from specialized object detection models.

Our contributions are as follows:
\begin{itemize}
    \item We provide a comprehensive analysis of sota VLMs in counting tasks, demonstrating that counting remains systematically the lowest-accuracy task across all evaluated models despite advances in reasoning capabilities.
    \item We introduce \textbf{GroundCount}, an object-detection-driven augmentation pipeline that increases VLM counting accuracy by 6.2 to 7.5pp across four of five evaluated models (up to 81.3\% on the PhD benchmark) with negligible memory overhead and reduced inference time for stronger models (Ovis2.5-2B: -22\%, Molmo2-4B: -23\%).
    \item We conduct three groups of ablation studies analyzing the impact of different ODM outputs (confidence scores, positional encoding, detection thresholds) on VLM performance, providing insights into why grounding information reduces hallucination.
    \item We evaluate a fusion architecture that combines VLMs with ODMs at the feature level through fine-tuning, analyzing the trade-offs between prompt-based augmentation and architectural integration.
\end{itemize}

\newcolumntype{S}{>{\centering\arraybackslash}m{0.35cm}}
\newcommand{\num}[1]{{\footnotesize #1}}
\newcommand{\smalltask}[1]{{\footnotesize #1}}
\newlength{\grpsep}      \setlength{\grpsep}{4pt}
\newlength{\thinrule}    \setlength{\thinrule}{0.15pt}
\definecolor{lightred}{RGB}{255,200,200}
\definecolor{lightorange}{RGB}{255,229,180}
\newcommand{\lowest}[1]{\cellcolor{lightred}{\footnotesize #1}}
\newcommand{\secondlowest}[1]{\cellcolor{lightorange}{\footnotesize #1}}
\begin{table*}[hb]
\centering
\caption{Correctness rates (\%) of different VLMs on the PhD benchmark using greedy decoding, with explicit thinking mode enabled when available (i.e., the output begins with a prior reasoning segment enclosed by special tokens such as \texttt{"<think>...</think>"}).
The PhD benchmark consists of questions from five task types listed in the \textbf{Task} column on the left.
For each subset (excluding sentiment), the task with the lowest correctness rate is highlighted with a \colorbox{lightred}{light-red} background.
Results across all four evaluation settings are shown; baseline corresponds to standard VQA without adversarial context.}
\label{tab:results_baseline}
\vspace{-0.1cm}
\begin{tabular}{c|%
    SSSS !{\hskip \grpsep\vrule width \thinrule\hskip \grpsep}%
    @{\hspace{3pt}}SSSS !{\hskip \grpsep\vrule width \thinrule\hskip \grpsep}%
    @{\hspace{3pt}}SSSS !{\hskip \grpsep\vrule width \thinrule\hskip \grpsep}%
    @{\hspace{3pt}}SSSS !{\hskip \grpsep\vrule width \thinrule\hskip \grpsep}%
    @{\hspace{3pt}}SSSS}
\hline
\multicolumn{1}{c|}{\textbf{VLM}}
& \multicolumn{4}{@{}c@{\hskip \grpsep\vrule width \thinrule\hskip \grpsep}}{\textbf{Molmo2-4B}}
& \multicolumn{4}{@{\hspace{3pt}}c@{\hskip \grpsep\vrule width \thinrule\hskip \grpsep}}{\textbf{Ovis2.5-2B}}
& \multicolumn{4}{@{\hspace{3pt}}c@{\hskip \grpsep\vrule width \thinrule\hskip \grpsep}}{\textbf{R-4B}}
& \multicolumn{4}{@{\hspace{3pt}}c@{\hskip \grpsep\vrule width \thinrule\hskip \grpsep}}{\textbf{Qwen3-VL-2B}}
& \multicolumn{4}{@{\hspace{3pt}}c@{}}{\textbf{InternVL3.5-1B}} \\
\cline{1-21}
\textbf{Task} & \smalltask{base} & \smalltask{sec} & \smalltask{icc} & \smalltask{ccs} 
& \smalltask{base} & \smalltask{sec} & \smalltask{icc} & \smalltask{ccs} 
& \smalltask{base} & \smalltask{sec} & \smalltask{icc} & \smalltask{ccs}
& \smalltask{base} & \smalltask{sec} & \smalltask{icc} & \smalltask{ccs}
& \smalltask{base} & \smalltask{sec} & \smalltask{icc} & \smalltask{ccs} \\
\hline
{\small object}     & {\small\num{89.6}} & {\small\num{79.2}} & {\small\num{65.9}} & {\small\num{89.2}} & {\small\num{88.5}} & {\small\num{85.1}} & {\small\num{84.0}} & {\small\num{84.6}} & {\small\num{87.9}} & {\small\num{85.3}} & {\small\num{82.1}} & {\small\num{90.4}} & {\small\num{87.7}} & {\small\num{84.3}} & {\small\num{83.1}} & \lowest{83.1} & {\small\num{70.3}} & {\small\num{68.3}} & {\small\num{67.9}} & {\small\num{82.0}} \\
{\small attribute}  & {\small\num{85.2}} & \lowest{44.7} & \lowest{15.1} & {\small\num{93.9}} & {\small\num{86.5}} & {\small\num{76.8}} & {\small\num{75.3}} & {\small\num{89.2}} & {\small\num{86.5}} & {\small\num{75.5}} & {\small\num{64.5}} & {\small\num{92.7}} & {\small\num{86.5}} & {\small\num{77.8}} & {\small\num{74.2}} & {\small\num{88.1}} & {\small\num{83.2}} & {\small\num{75.2}} & {\small\num{73.6}} & \lowest{80.9} \\
{\small positional} & {\small\num{80.8}} & {\small\num{66.4}} & {\small\num{47.3}} & {\small\num{89.1}} & {\small\num{80.9}} & {\small\num{73.0}} & {\small\num{65.3}} & {\small\num{88.6}} & {\small\num{80.6}} & {\small\num{71.5}} & {\small\num{54.5}} & {\small\num{94.1}} & {\small\num{78.3}} & {\small\num{71.3}} & {\small\num{63.6}} & {\small\num{86.8}} & {\small\num{70.2}} & {\small\num{61.3}} & {\small\num{49.2}} & {\small\num{85.9}} \\
{\small counting}   & \lowest{67.4} & {\small\num{62.9}} & {\small\num{33.1}} & \lowest{87.9} & \lowest{74.7} & \lowest{67.6} & \lowest{52.0} & \lowest{79.8} & \lowest{73.7} & \lowest{69.4} & \lowest{44.2} & \lowest{87.9} & \lowest{69.4} & \lowest{63.5} & \lowest{53.3} & \lowest{83.1} & \lowest{64.0} & \lowest{54.0} & \lowest{41.3} & {\small\num{85.5}} \\
\cdashline{1-21}
{\small sentiment}  & {\small\num{67.3}} & {\small\num{52.9}} & {\small\num{32.0}} & {\small\num{93.6}} & {\small\num{68.4}} & {\small\num{59.6}} & {\small\num{59.2}} & {\small\num{94.9}} & {\small\num{68.1}} & {\small\num{57.9}} & {\small\num{50.0}} & {\small\num{96.2}} & {\small\num{68.1}} & {\small\num{50.9}} & {\small\num{50.3}} & {\small\num{91.0}} & {\small\num{66.2}} & {\small\num{57.7}} & {\small\num{59.7}} & {\small\num{83.3}} \\
\hline
\end{tabular}
\end{table*}

\section{Related Works}
\label{sec:related_works}

\textbf{VLM Hallucination in Counting Tasks}: 
Vision-Language Models exhibit persistent hallucinations in counting tasks despite advances 
in reasoning capabilities. One major hypothesis attributes this to vision transformers (ViTs) 
having difficulties noticing fine-grained image details \cite{park2025second, raghu2021vit}, 
as ViTs process images through global patch attention rather than the local receptive fields 
that facilitate spatial instance discrimination \cite{naseer2021intriguing, park2025second}.
Recent studies~\cite{sengupta2025visionlanguagemodelscountsynthetic, vo2025vision} 
systematically demonstrate that counting remains a fundamental challenge for VLMs, with 
accuracy substantially lower than other visual reasoning tasks. 
The autoregressive nature of VLM decoding may also contribute to counting errors, as the 
model must maintain accurate object tallies across many generation steps without explicit 
grounding in visual features.

\textbf{VLM Hallucination Mitigation Methods}: 
Existing approaches to mitigate VLM hallucinations operate at multiple levels. Decoding-level 
methods adjust the generation process through contrastive decoding \cite{yang2025mitigating}, 
layer contrasting \cite{chuang2024dola}, or attention-guided decoding \cite{yin2025clearsight}. 
Training-level approaches incorporate regularization \cite{wu2025antidote} or specialized 
fine-tuning objectives. More recent steering strategies \cite{wang2025damo, wang2025mllm} 
preserve correct activations from earlier layers to prevent hallucination drift. However, these 
methods primarily target general hallucinations and show limited effectiveness on systematic 
compositional failures like counting, particularly in reasoning-capable VLMs that already 
employ reflection mechanisms.

\textbf{Combining CNN with Transformers}: 
The complementary strengths of CNNs and transformers have motivated hybrid architectures. 
DETR \cite{carion2020detr} pioneered end-to-end object detection by combining CNN backbones 
with transformer decoders, demonstrating how structured CNN features can enhance transformer 
reasoning. More recent work explores various fusion strategies for integrating convolutional 
inductive biases with transformer flexibility \cite{liu2022convnet, dai2021coatnet}. 
Our fusion architecture extends this paradigm to multimodal settings by grounding VLM patch 
tokens with CNN-based object detection features.

\textbf{Fusion Training and Cross-Modal Integration}: 
Training multimodal fusion architectures requires careful alignment of features from different 
modalities and architectural paradigms \cite{lu2019vilbert, tan2019lxmert}. 
Cross-attention mechanisms have proven effective for selectively integrating information across 
modalities~\cite{carion2020detr, alayrac2022flamingo, li2023blip2}.
Feature-wise Linear Modulation (FiLM)~\cite{perez2018film} provides adaptive conditioning 
mechanisms that allow one modality to modulate another's representations. 
Recent work on vision-language pre-training emphasizes joint optimization of vision encoders 
and language decoders \cite{zhu2025internvl3}, though these typically focus on 
transformer-only architectures. Our fusion approach specifically targets integrating CNN-based 
object detection with transformer-based VLMs, requiring novel architectural components to 
bridge the representational gap between local CNN features and global ViT patch embeddings.

\textbf{Object Detection for Visual Reasoning}: 
Object detection models provide structured, localized visual information that can ground 
higher-level reasoning. YOLO \cite{Redmon2016YOLO} and its variants achieve real-time 
detection with high accuracy through efficient CNN architectures. These models excel at spatial 
localization and instance counting, which are precisely the capabilities where VLMs struggle. 
Prior work has explored using detected objects as input to VQA systems 
\cite{hudson2019gqa, li2020bertvision}, but typically through learned feature fusion during 
training rather than explicit textual grounding \cite{anderson2018bottomup, chen2020uniter}. 
Our prompt-based approach (Plan A) uniquely leverages ODM outputs as interpretable, structured 
prompts that augment VLMs without architectural modification, while Plans B and C extend this 
with feature-level integration for deeper cross-modal grounding.

\section{Analysis - Hallucination in Counting Tasks}
\label{sec:analysis}

\begin{figure*}[hb]
    \centering
    \includegraphics[width=\textwidth, trim={5 30 65 35}, clip]{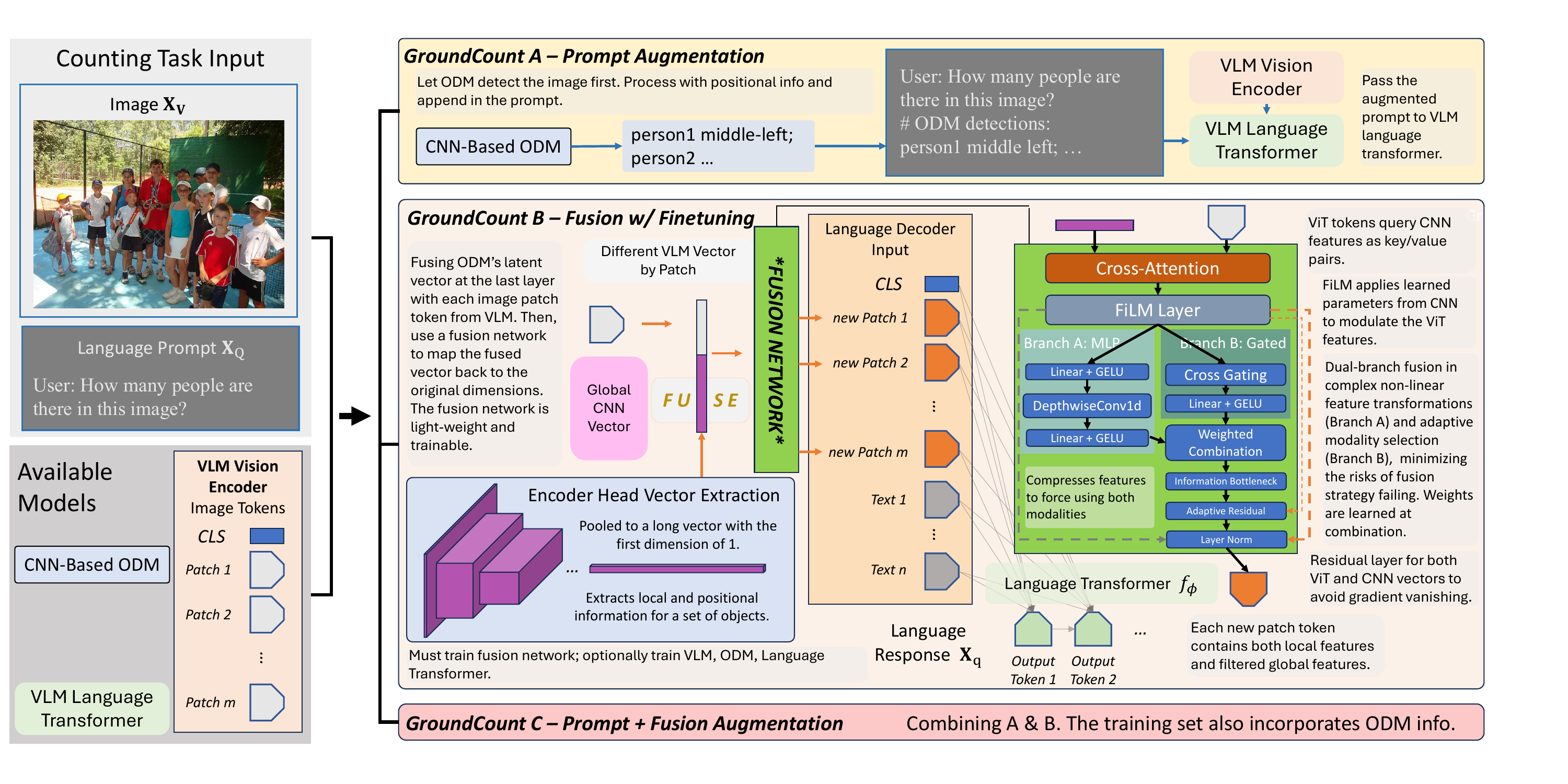}
    \vspace{-0.5cm}
    \caption{Structural overview of three strategies in our proposed fusion framework - A, B and C. In \textbf{GroundCount A}, we run inference with ODM on the image, and then include its output in the VLM prompt. In \textbf{GroundCount B}, we fuse the VLM and ODM on the visual patch latent vector using a light-weight network. To ensure correct information delivery, we finetune the network with our original counting task mutation from COCO. The fusion block is required to be trained; Other modules - VLM, ODM, and language transformer - are optionally frozen. \textbf{GroundCount C} incorporates both plans by including both prompt-level information and architectural-level integration. The training data also includes ODM detections in the textual input.}
    \label{fig:full_methodology}
\end{figure*}

We first validate that state-of-the-art VLMs systematically exhibit counting hallucinations through controlled experiments. We employ consistent benchmark and model selections across all evaluations in this work.

\subsection{VQA Benchmark}

We employ the PhD benchmark~\cite{liu2025phd} to evaluate vision-language model capabilities. The benchmark comprises 33,688 visual question-answer (VQA) pairs across 16,844 images with binary (yes/no) ground-truth labels. Questions are categorized into five task types: \textit{attribute}, \textit{counting}, \textit{object}, \textit{positional}, and \textit{sentiment}. 
To assess robustness against misleading information, the benchmark provides three challenge variants: (1) \textit{sec} (specious context) with plausible but potentially misleading descriptions, (2) \textit{icc} (incorrect context) where VLMs receive explicitly incorrect textual descriptions, and (3) \textit{ccs} (contradictory common sense) featuring 1,506 VQA pairs on 753 AI-generated images that violate common-sense expectations. Compared to earlier benchmarks such as POPE~\cite{pope2023li} and CHAIR~\cite{chair2018rohrbach}, PhD provides larger sample sizes and more granular task categorization.

\subsection{Selection of VLM and ODM}

We evaluate five state-of-the-art open-source VLMs with parameters not exceeding 4B: R-4B~\cite{yang2025r4b}, Ovis2.5-2B~\cite{lu2025ovis25}, Qwen3-VL-2B-Thinking (abbreviated as Qwen3-VL-2B for the rest of the paper) ~\cite{bai2025qwen3vltechnicalreport}, InternVL3.5-1B~\cite{wang2025internvl35}, and Molmo2-4B~\cite{clark2026molmo2openweightsdata}. This selection spans diverse architectural designs, with several incorporating advanced reasoning mechanisms such as reinforcement learning (R-4B) and iterative reflection (InternVL3.5). All models are loaded in \texttt{float32} precision. We allocate generous computational budgets of 1,024 tokens for both output generation and thinking steps, using greedy decoding to ensure reproducibility.
For object detection, we utilize YOLOv13x~\cite{lei2025yolov13} as our primary ODM, selected for its state-of-the-art accuracy and minimal inference latency.

\subsection{Results and Discussion}

Table~\ref{tab:results_baseline} presents the comprehensive evaluation results. Our findings clearly demonstrate: \textbf{counting consistently remains the lowest-accuracy task in standard evaluation, with attribute identification often showing steeper degradation under adversarial contexts in higher-capacity models} (other than sentiment, which is relatively subjective).

In the base evaluation setting, counting accuracy ranges from 64.0\% to 74.7\% (mean: 69.8\%), representing a substantial gap compared to other visual reasoning tasks: object recognition achieves 70.3\%--89.6\% (mean: 84.8\%), attribute identification reaches 83.2\%--86.5\% (mean: 85.6\%), and positional reasoning attains 70.2\%--80.9\% (mean: 78.2\%). This represents an average accuracy deficit of 15.0pp and 15.8pp compared to object recognition and attribute identification, respectively.

Performance degradation under misleading contexts further reveals the fragility of counting abilities. Under \textit{sec} and \textit{icc} conditions, counting accuracy drops by 4.3--10.0pp and 16.1--34.3pp, respectively, the steepest degradation among all task categories. This suggests that VLM counting mechanisms are particularly susceptible to distraction from textual priors, consistent with hypotheses about imbalanced cross-modal attention~\cite{Leng_2024_CVPR}.

Interestingly, the \textit{ccs} subset reveals a reversal: VLMs achieve substantially higher counting accuracy (79.8\%--87.9\%) on AI-generated images. We hypothesize this occurs due to cleaner spatial layouts and reduced perceptual ambiguity in synthetic images, though this improved performance does not reflect real-world applicability.

\textbf{Model-Specific Patterns.}
Examining individual models reveals that counting failures are not simply correlated with parameter count or training paradigm. Molmo2-4B achieves only 67.4\% baseline accuracy despite having the largest parameter count among evaluated models (tied with R-4B at 4B), while the smaller Ovis2.5-2B reaches 74.7\%. However, Molmo2 demonstrates the second smallest accuracy drop under misleading contexts (4.5pp for \textit{sec}, second to 4.3pp for \textit{sec} of R-4B), suggesting robust attention mechanisms. InternVL3.5-1B exhibits the steepest degradation under \textit{sec} (10.0pp) in the counting task, indicating that iterative reflection mechanisms require sufficient capacity to filter spurious textual information.

These findings have critical implications. Despite incorporating advanced reasoning capabilities, modern VLMs continue to exhibit systematic counting failures. This suggests the problem lies not in reasoning depth but in fundamental spatial-semantic integration. Moreover, models explicitly addressing cross-modal attention balance still achieve only 64.0\% to 74.7\% counting accuracy, indicating that attention rebalancing strategies have limited impact. Layer-level steering strategies~\cite{wang2025damo} prove ineffective for iterative reasoning processes where no single "correct" token exists to steer toward during multi-step counting.

These observations motivate our approach: rather than refining attention mechanisms within existing VLM paradigms, we augment VLMs with explicit grounding from specialized object detection models that excel precisely where VLMs fail.

\section{Methodology - Augmenting VLMs with Object Detection Model}
\label{sec:methodology}

We propose GroundCount, a framework that augments VLMs with explicit spatial grounding from object detection models to mitigate counting hallucinations. Our approach operates on the observation that CNN-based ODMs excel at spatial localization and instance counting, precisely where VLMs exhibit systematic failures. We present three implementation strategies with different computational trade-offs.

\subsection{GroundCount A: Prompt-Based ODM Augmentation}
\label{sec:method_a}

\begin{figure}[ht]
    \centering
    \includegraphics[width=\columnwidth, trim={37 120 1070 45}, clip]{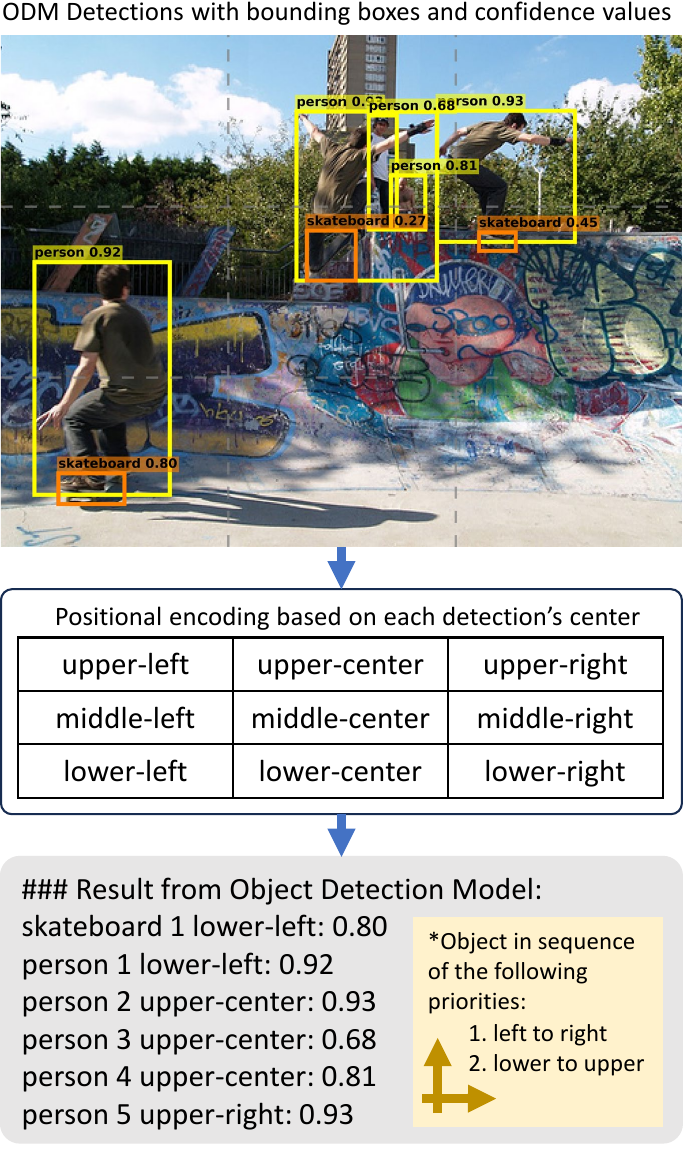}
    \vspace{-0.5cm}
    \caption{Our pipeline of converting ODM outputs to descriptive text. The image is \#000000000077.jpg from COCO-train2017, showing 5 young people skateboarding. 
    The bounding boxes (bbox) come from YOLOv13x's detection: yellow ones are \texttt{person} objects; orange ones are \texttt{skateboard} objects. The location of each object is determined by the center of their corresponding bbox. Two skateboard objects were not included due to low confidence.}
    \label{fig:odm_to_string}
\end{figure}

The most straightforward approach augments VLM prompts with structured textual descriptions of ODM detections. As illustrated in Figure~\ref{fig:odm_to_string}, we first pass the input image through YOLOv13x to obtain bounding boxes, class labels, and confidence scores, which are then converted into natural language prompts.

\textbf{Spatial Encoding:} We discretize image space into a 3×3 grid (upper/middle/lower × left/center/right) and assign each detection to a grid cell based on its bounding box center coordinates. This coarse spatial encoding preserves relative positioning information while remaining interpretable for language models.

\textbf{Object Sequencing:} Detections are ordered by (1) horizontal position (left-to-right), then (2) vertical position (lower-to-upper). This consistent ordering enables VLMs to maintain spatial coherence when processing multiple instances of the same object class.

\textbf{Prompt Construction:} For each detection, we generate a string in the format: \texttt{"[class] [index] [position]: [confidence]"}, where the index distinguishes multiple instances of the same class. The complete ODM output is appended to the original user prompt as structured context (see Figure~\ref{fig:odm_to_string}).

\begin{figure*}[ht]
    \centering
    \includegraphics[width=\textwidth, trim={10 75 0 0}, clip]{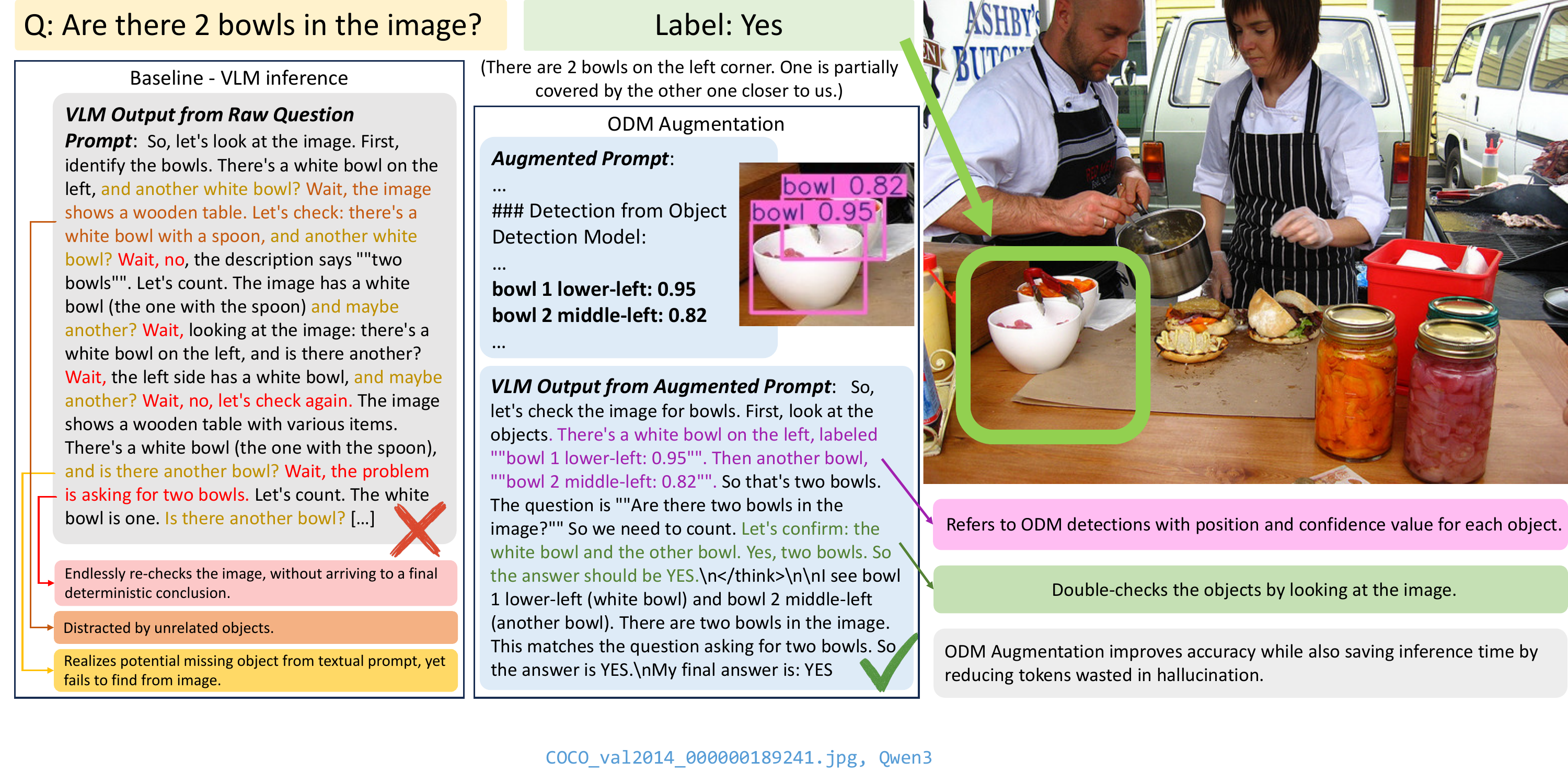}
    \vspace{-0.5cm}
    \caption{Illustration with real evaluation example for ODM prompt augmentation in GroundCount A. The image is \#000000189241.jpg from COCO-val2014, which is used for a counting question in our selected benchmark, PhD. The tested VLM is Qwen3-VL-2B-Thinking. The question asks for a correctness judgment on the number of bowls in the image. There are indeed 2 bowls in the image, with one bowl partially covered by another. The baseline VLM, whose output is showcased in the left frame, fails to find the second bowl and re-thinks iteratively, exhibiting behaviors of hallucination. On the other hand, with information from the object detection model (ODM) appended in the prompt, the VLM successfully finds the second bowl with a double-check.}
    \label{fig:plan_a}
\end{figure*}

This approach requires no architectural modifications or training, enabling plug-and-play augmentation across different VLM families. The computational overhead is minimal: YOLOv13x inference ($\sim$0.1s) is negligible compared to VLM auto-regressive decoding (7-40s depending on model architecture and reasoning complexity), and sometimes reduces total inference time by preventing hallucination-induced reasoning loops, particularly for stronger models (Figure~\ref{fig:plan_a}).

\subsection{GroundCount B: Feature-Level Fusion Architecture}
\label{sec:method_b}

While prompt augmentation proves effective, it relies on the VLM's ability to correctly interpret textual descriptions of spatial information. To enable direct feature-level grounding, we propose a fusion architecture that integrates CNN features from the ODM with ViT patch tokens from the VLM's vision encoder.

\textbf{Architecture Overview:} As shown in Figure~\ref{fig:full_methodology}, our fusion network operates between the VLM's vision encoder and language decoder. For each ViT patch embedding $\mathbf{p}_i \in \mathbb{R}^{d_{\text{vit}}}$, we extract the corresponding spatial region from the ODM's final convolutional layer, yielding a local CNN feature $\mathbf{c}_i \in \mathbb{R}^{d_{\text{cnn}}}$. We additionally extract a global CNN feature vector $\mathbf{g} \in \mathbb{R}^{d_{\text{cnn}}}$ via adaptive pooling.

\textbf{Dual-Branch Fusion:} The fusion network employs two parallel branches to integrate multimodal features:

\textit{Branch A (Feature Transformation):} Applies Feature-wise Linear Modulation (FiLM)~\cite{perez2018film} to enable CNN features to adaptively modulate ViT representations:
$$\mathbf{h}_i^A = \text{FiLM}(\mathbf{p}_i, \mathbf{c}_i) = \boldsymbol{\gamma}_i \odot \mathbf{p}_i + \boldsymbol{\beta}_i$$
where $\boldsymbol{\gamma}_i, \boldsymbol{\beta}_i = \text{MLP}(\mathbf{c}_i)$ are learned affine parameters.

\textit{Branch B (Selective Attention):} Uses cross-attention to allow ViT patches to selectively query relevant CNN features:
$$\mathbf{h}_i^B = \text{CrossAttn}(\mathbf{p}_i, [\mathbf{c}_i, \mathbf{g}])$$

The branches are combined via learned gating: $\mathbf{h}_i = \alpha_i \mathbf{h}_i^A + (1-\alpha_i) \mathbf{h}_i^B$, where $\alpha_i = \sigma(\text{MLP}([\mathbf{p}_i, \mathbf{c}_i]))$. This dual-branch design provides robustness. Branch A enforces strong CNN influence through multiplicative modulation, while Branch B enables adaptive selection. This minimizes risk if one fusion strategy proves suboptimal.

\textbf{Information Bottleneck:} Following the dual-branch fusion, we apply dimensionality reduction to enforce multimodal information integration. This bottleneck prevents the model from trivially bypassing fusion by relying solely on one modality.

\textbf{Training Data Preparation:} We construct training data from the COCO train2017 dataset \cite{lin2014microsoft} using ground-truth object annotations. For each image, we apply the same spatial encoding and sequencing rules used for ODM detections in Plan A. Specifically, each ground-truth bounding box is assigned to a 3×3 grid cell based on its center coordinates, and objects are ordered left-to-right, then lower-to-upper.

The model is trained to generate structured spatial descriptions in the format: \texttt{"[class] [index] in [position]"} for each object instance. For example, given an image with multiple birds, the target output would be: \textit{"bird 1 in upper-left; bird 2 in middle-center; bird 3 in lower-right"}. This format mirrors the ODM detection output structure, enabling the fusion model to learn consistent spatial grounding patterns that align with both ground-truth annotations and runtime ODM predictions.

\textbf{Training Strategies:} We evaluate four training configurations (Table~\ref{tab:results_training}):
\begin{itemize}
\item \textbf{B.1}: Train fusion network only (frozen VLM, frozen ODM)
\item \textbf{B.2}: Train fusion network + fine-tune VLM and ODM
\item \textbf{B.3}: Train fusion network + fine-tune language decoder only
\item \textbf{B.4}: Train fusion network + fine-tune VLM, ODM, and language decoder
\end{itemize}

\subsection{GroundCount C: Combined Prompt and Fusion}
\label{sec:method_c}

Our final approach combines both strategies: structural fusion (Plan B.4) with prompt augmentation (Plan A). This hybrid design provides complementary benefits. Feature-level fusion enables implicit spatial grounding, while textual prompts offer explicit, interpretable object counts. The combined approach aims to leverage both implicit feature integration and explicit symbolic reasoning.

\subsection{Implementation Details}

All experiments use YOLOv13x as the object detection model with default confidence threshold 0.5. For Plan A, we filter detections below this threshold before prompt construction. VLMs are evaluated with greedy decoding, 1024-token output budget, and float32 precision. Fusion network training uses AdamW optimizer ($\beta_1=0.9, \beta_2=0.999$), learning rate $2 \times 10^{-5}$ with cosine annealing, batch size 1, and training with a maximum training budget of 40k steps; best checkpoints selected per Table \ref{tab:results_training}. All experiments are conducted on NVIDIA A100 GPUs with 80GB memory.

\section{Results \& Analysis}
\label{sec:results}

\subsection{Main Results: Comparing GroundCount Strategies}

Table~\ref{tab:results_training} presents a comprehensive comparison of our proposed GroundCount strategies on the PhD benchmark's counting subset, evaluated on Ovis2.5-2B.

\textbf{Prompt Augmentation (Plan A)} achieves the highest counting accuracy at 81.3\%, representing a substantial 6.6pp improvement over the 74.7\% baseline. Remarkably, Plan A also reduces average inference time from 10.0s to 7.8s (a 22\% speedup). This counterintuitive result stems from VLMs generating fewer tokens when provided with explicit grounding information (Figure~\ref{fig:plan_a}). The negligible computational overhead of YOLOv13x inference ($\sim$0.1s) is more than offset by reduced VLM generation, making Plan A both more accurate and faster. This efficiency gain is particularly notable in complex reasoning scenarios where VLMs typically engage in extended chain-of-thought generation.

\textbf{Fusion Architectures (Plan B)} demonstrate varied outcomes depending on training configuration. Training only the fusion network (B.1) yields minimal improvement (75.2\%), suggesting that frozen VLM and ODM representations remain misaligned. Conversely, jointly fine-tuning the fusion network, VLM, and ODM (B.2) degrades performance to 72.7\%, indicating potential catastrophic forgetting. The most effective fusion strategy (B.4) fine-tunes the fusion network, VLM, ODM, and language decoder together, achieving 78.0\% accuracy with 7.7s inference time. While this represents a meaningful 3.3pp gain over baseline, it falls short of Plan A's prompt-based approach.

\textbf{Combined Strategy (Plan C)} integrates prompt augmentation with fusion architecture (B.4), yielding 78.2\% accuracy and 4.8s inference (the fastest among approaches that exceed baseline accuracy). The reduced inference time stems from the fine-tuned language decoder generating more concise responses. However, the accuracy remains below Plan A alone, suggesting that the fusion network may introduce noise that partially counteracts the benefits of explicit textual grounding.

These findings reveal a critical insight: for counting tasks, \emph{explicit symbolic grounding via prompts outperforms implicit feature-level fusion}. This aligns with recent work emphasizing the importance of interpretable intermediate representations in multimodal reasoning~\cite{li2023blip2}. The fusion architecture's underperformance likely stems from the fundamental representational gap between CNN spatial features and ViT global patch embeddings.

\begin{figure*}[ht]
    \centering
    \includegraphics[width=\textwidth, trim={5 0 5 0}, clip]{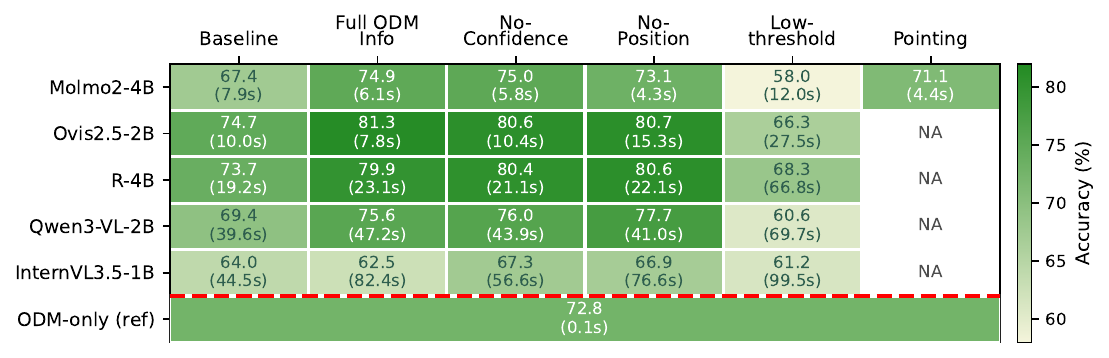}
    \vspace{-0.7cm}
    \caption{Results of GroundCount A across all model families and including ablation studies. Each block contains the accuracy and average inference time for that group of experiment on the PhD counting subset. The bottom row marks the result of running the object detection model only. The right-most column records the special pointing mode for Molmo2 model only.}
    \label{fig:major_heatmap}
\end{figure*}

\subsection{Ablation Study: Decomposing ODM Information}

Figure~\ref{fig:major_heatmap} presents ablation results across five VLM families, systematically removing different components of ODM information.

\textbf{Confidence Scores}: Removing confidence values (NoConfidence) produces mixed results. While Ovis2.5-2B shows a modest 0.7pp accuracy drop, four models (Molmo2-4B, R-4B, Qwen3-VL-2B, InternVL3.5-1B) actually improve by 0.1 to 4.8pp without confidence scores, suggesting that confidence values may introduce noise for certain VLM architectures.

\textbf{Positional Encoding}: The impact of spatial position information reveals a critical architectural divide. Removing positional encoding (NoPosition) causes degradation in the two strongest baseline models: Molmo2-4B (1.8pp loss) and Ovis2.5-2B (0.6pp loss). However, the three weaker models—R-4B, Qwen3-VL-2B, and InternVL3.5-1B—paradoxically \emph{improve} by 0.7 to 4.4pp without positional information, suggesting these models struggle to correctly interpret spatial encodings and perform better with position-agnostic object lists.

\textbf{Detection Threshold}: Lowering the confidence threshold from 0.5 to 0.3 (LowThreshold) uniformly harms performance across all models, with accuracy decreasing by 1.3 to 16.9pp relative to the Full ODM augmentation, and inference time increasing by 4.1 to 55.0s relative to the no-augmentation baseline. The degradation is particularly severe for stronger models (Molmo2-4B: 16.9pp, Ovis2.5-2B: 15.0pp). The dramatic slowdown occurs because VLMs must process substantially more false-positive detections. This establishes that \emph{detection precision substantially outweighs recall} for counting augmentation.

\textbf{Model-Specific Patterns}: Molmo2-4B uniquely supports a pointing mode where bounding boxes are overlaid on the input image. Interestingly, this visual grounding (71.1\%) underperforms textual ODM augmentation (74.9\%), suggesting that Molmo's architecture processes structured textual descriptions more effectively than visual annotations.

\subsection{ODM-Only Baseline}

Running YOLOv13x alone achieves 72.8\% accuracy in 0.1s. This establishes an important reference point: the ODM already captures most counting information, yet VLMs struggle to extract it from raw images. When properly grounded with ODM outputs, the strongest VLMs (Plan A: Ovis2.5-2B at 81.3\%) exceed ODM-only performance by 8.5pp, demonstrating that VLMs contribute valuable contextual reasoning when supplied with explicit spatial priors.

\subsection{Cross-Model Consistency and Architectural Variations}

The benefits of ODM augmentation show significant variation across VLM architectures. Four of five evaluated models exhibit substantial accuracy improvements with Plan A: Molmo2-4B (7.5pp), Ovis2.5-2B (6.6pp), R-4B (6.2pp), and Qwen3-VL-2B (6.2pp).

However, InternVL3.5-1B presents a notable exception, showing degraded performance (from 64.0\% baseline to 62.5\% with full ODM augmentation). This reveals that InternVL3.5's architecture appears incompatible with explicit textual grounding, potentially due to its iterative reflection mechanisms being disrupted by structured prompts. The model's improvements in NoPosition (+4.4pp) and NoConfidence (+4.8pp) ablations suggest it performs better with minimal explicit guidance.

This divergence suggests that while counting hallucinations generally stem from fundamental limitations in spatial-semantic integration, the effectiveness of explicit grounding depends critically on architectural compatibility.

\subsection{Implications and Future Directions}

Our results establish that counting hallucinations in VLMs reflect deeper challenges in spatial-semantic integration. The success of prompt-based augmentation over sophisticated fusion architectures suggests that current VLMs are better equipped to process structured symbolic representations than to learn implicit cross-modal alignments. However, the architectural variation—particularly InternVL3.5's negative response—indicates that augmentation strategies must be tailored to specific VLM designs.

Several limitations warrant future investigation. First, our fusion experiments used only 40k training steps on binary classification tasks. More extensive pre-training may unlock greater potential. Second, we focused on CNN-based detectors (YOLO). Exploring transformer-based ODMs like DETR or Grounding DINO could reduce the representational gap. Third, understanding why certain architectures reject explicit grounding could inform better augmentation strategies. Finally, efficient caching for repeated queries would enhance practical applicability.

\begin{table}[ht]
\centering
\caption{Performance comparison of different GroundCount schemes - prompt augmentation (Plan A), architectural fusion with training (Plan B), and both (Plan C). Plan A gains the highest accuracy; Plan B.4 and Plan C have the lowest inference time among schemes that exceed baseline accuracy, due to training on language transformer.}
\label{tab:results_training}
\vspace{-0.1cm}
\begin{tabular}{p{4.5cm}>{\centering\arraybackslash}p{0.7cm}>{\centering\arraybackslash}p{0.7cm}>{\centering\arraybackslash}p{1.3cm}}
\toprule
\textbf{Training Scheme} & \textbf{Acc(\%)} & \textbf{Time(s)} & \textbf{Best Steps} \\
\midrule
Baseline & 74.7 & 10.0 & NA \\
Plan A - ODM Prompt Augmentation & 81.3 & 7.8 & NA \\
Plan B.1 - Fusion only & 75.2 & 17.8 & 10k \\
Plan B.2 - Fusion + VLM, ODM & 72.7 & 14.6 & 5k \\ 
Plan B.3 - Fusion + LangTrans & 71.4 & 4.3 & 5k \\
Plan B.4 - Fusion + VLM, ODM & \multirow{2}{*}{78.0} & \multirow{2}{*}{7.7} & \multirow{2}{*}{5k} \\
+ LangTrans & & & \\
Plan C - Plan A + Plan B.4 & 78.2 & 4.8 & 5k \\
\bottomrule
\end{tabular}
\end{table}

\section{Conclusion}
\label{sec:conclusion}
We present GroundCount, a framework that mitigates counting hallucinations in VLMs through explicit grounding from object detection models. Our evaluation across five state-of-the-art VLMs shows counting remains the lowest-accuracy task (64.0 to 74.7\%, excluding sentiment). 
Our prompt-based augmentation (Plan A) achieves 81.3\% counting accuracy—a 6.6pp improvement over baseline—while reducing inference time by 22\%. Ablation studies reveal positional encoding is beneficial for stronger models (0.6 to 1.8pp), while detection precision substantially outweighs recall (1.3 to 16.9pp degradation at lower thresholds).
Critically, we find that explicit symbolic grounding via structured prompts outperforms feature-level fusion on Ovis2.5-2B, while prompt-based augmentation achieves consistent gains across four of five architectures. This indicates counting failures stem from fundamental spatial-semantic integration limitations rather than architecture-specific deficiencies. The gap between ODM-only (72.8\%) and augmented VLM performance (81.3\%) demonstrates that VLMs contribute valuable contextual reasoning when properly grounded.

\section*{Impact Statement}
This work improves VLM reliability in counting tasks, enhancing trustworthiness for accessibility tools, inventory systems, and educational technologies. We acknowledge that improved counting accuracy could amplify privacy concerns if applied to surveillance without appropriate safeguards. We encourage responsible deployment with proper consent and adherence to privacy regulations. Reducing hallucinations contributes to developing more reliable multimodal AI systems.

\section*{Acknowledgments}
This work has been supported in parts by the NYUAD Center for Cyber Security (CCS), funded by Tamkeen under the NYUAD Research Institute Award G1104. Experiments are performed with NYUAD Jubail High Performance Computing (HPC). 

\bibliographystyle{IEEEtran}
\bibliography{reference}


\end{document}